\documentclass[journal]{IEEEtran}
\usepackage{times}
\usepackage{epsfig}
\usepackage{graphicx}
\usepackage{amsmath}
\usepackage{amssymb}
\usepackage[pagebackref=true,breaklinks=true,letterpaper=true,colorlinks,bookmarks=false]{hyperref}

\usepackage{subfigure}
\usepackage{float}
\usepackage{amsmath}

\usepackage{algorithm}
\usepackage{algorithmic}
\usepackage{mdwlist}
\usepackage{multirow}
\usepackage{bigstrut}

\usepackage{url}

\DeclareMathOperator*{\argmin}{\arg\!\min}
\DeclareMathOperator*{\argmax}{\arg\!\max}

\ifCLASSINFOpdf
\else
\fi
\begin{document}

\title{Query-free Clothing Retrieval via Implicit Relevance Feedback}
\author{Zhuoxiang~Chen,
        Zhe~Xu,
        Ya~Zhang,~\IEEEmembership{Member,~IEEE},
        and Xiao~Gu
\thanks{Zhuoxiang~Chen, Zhe~Xu, and Ya~Zhang are with Cooperative Medianet Innovation Center, Shanghai Jiao Tong University  (e-mail: \{chenzhuoxiang, xz3030, ya\_zhang\}@sjtu.edu.cn). Xiao~Gu is with Institute of Image Communication and Network Engineering, Shanghai Jiao Tong University  (e-mail: gugu97@sjtu.edu.cn). Ya Zhang is the corresponding author.}}
% <-this % stops a space


\maketitle

\begin{abstract}
Image-based clothing retrieval is receiving increasing interest with the growth of online shopping. In practice, users may often have a desired piece of clothing in mind (\textit{e.g.}, either having seen it before on the street or requiring certain specific clothing attributes) but may be unable to supply an image as a query. We model this problem as a new type of image retrieval task in which the target image resides only in the user's mind (called ``mental image retrieval'' hereafter). Because of the absence of an explicit query image, we propose to solve this problem through relevance feedback. Specifically, a new Bayesian formulation is proposed that simultaneously models the retrieval target and its high-level representation in the mind of the user (called the ``user metric'' hereafter) as posterior distributions of pre-fetched shop images and heterogeneous features extracted from multiple clothing attributes, respectively. Requiring only clicks as user feedback, the proposed algorithm is able to account for the variability in human decision-making. Experiments with real users demonstrate the effectiveness of the proposed algorithm.
\end{abstract}


\IEEEpeerreviewmaketitle

\section{Introduction}\label{sec::intro}

\IEEEPARstart{W}{e} have witnessed a dramatic upsurge in online shopping over the past few years. Among various kinds of products, a large portion of online purchasing activity is focused on clothing items. As a result, an increasing amount of research interest has been directed toward clothing item analysis, including attribute prediction \cite{abdulnabi2015multi,chen2015deep} and image retrieval \cite{yamaguchi2013paper}, with considerable success reported in the literature. %\cite{wang2013personal,jagadeesh2014large,vittayakorn2015runway,lin2015rapid}.

\begin{figure}[!t]
\centering
\includegraphics[width=.48\textwidth]{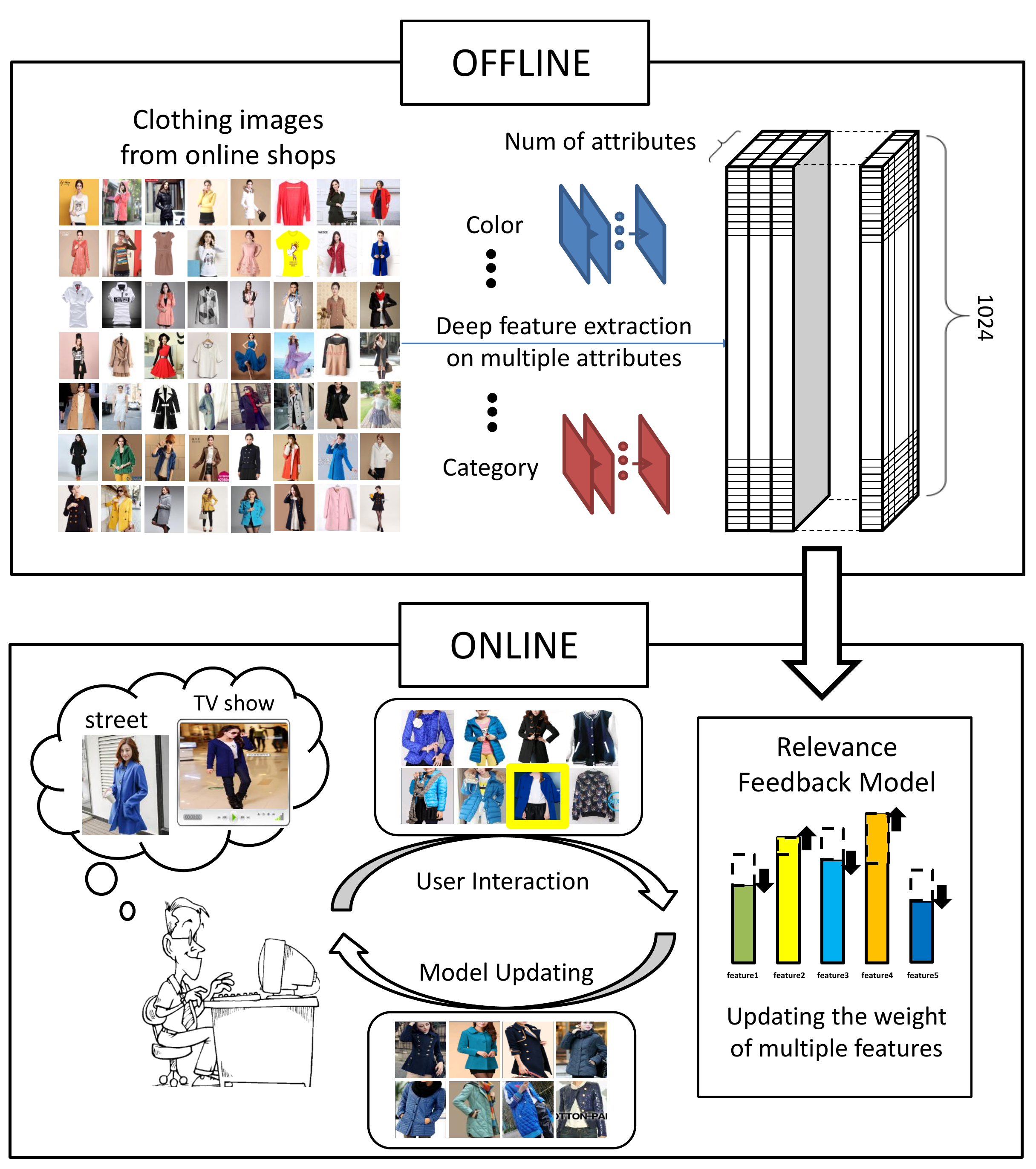}
\caption{Illustration of the proposed method. We address the query-free retrieval problem for clothing items, in which the target image resides only in the user's mind. Our method involves an offline process of extracting multiple deep convolutional features based on clothing attributes for images on online shopping websites and an online process of modeling the user's target image and selection criteria using implicit relevance feedback.}
\label{fig::title}
\end{figure}

%While traditional keyword-based retrieving scheme is still the dominant service provided in online shopping websites such as Amazon.com and eBay.com, recently researchers have begun to consider more interesting and practical shopping models involving content-based image retrieval. For example, beginning with a real-world photo of a clothing item taken on the street as the query, several methods are proposed to find the exact corresponding item \cite{hadi2015buy} or similar ones \cite{liu2012street,huang2015cross} in online shops by analyzing the visual content and attributes of clothing images. These methods, to some extent, solve the problem when an accurate keyword description is unavailable.

%Algorithms include the traditional keyword-based clothing retrieval and the more interesting and practical shopping models involving content-based image retrieval have both seen extensive use in online shopping websites such as Amazon.com and eBay.com.

In particular, several recent works \cite{wang2013personal,liu2012street,huang2015cross,hadi2015buy} have focused on image-content-based clothing retrieval, which avoids the often ambiguous keyword descriptions used in traditional retrieval scenarios. However, this process usually requires an explicit query image as input, which is a requirement that, unfortunately, cannot always be satisfied in real-world scenarios.
Imagine that you see a piece of clothing that you like on the street or in a TV show. In most cases, you may be unable or simply too polite to take a picture of it. Consequently, an image of this clothing item will reside only in your mind, and most likely, you will not have an opportunity to see it again, nor will you be able to find your desired product in online shops.% Being such a bad experience, what can we do to prevent this from happening?

To address the above scenario, we here investigate the novel application of query-free clothing retrieval, in which an image of the target item resides only in the user's mind. Because of the absence of an explicit query image, we propose to solve this problem via implicit relevance feedback \cite{zhou2003relevance}. Our method is inspired by the mental image retrieval algorithm \cite{fang2005experiments,ferecatu2007interactive,suditu2011heat,suditu2012iterative}, which introduced an interactive search paradigm based on user feedback. In this method, the target image is modeled as a random variable, and a cumulative probability distribution of this variable is stored to track past interactions with the user. This approach can be combined with textual description by asking the user to first provide several generic attributes as inputs for filtering (\textit{e.g.}, “red outerwear”). In this context, there are many ambiguous textual descriptions, such as “navy blue” or “bloomers”, that are unlikely to be accurately understood by an arbitrary user. Therefore, the method models the remaining “ambiguous” attributes via relevance feedback. In each iteration, the method selects a set of candidate images, shows them to the user, and then updates the underlying model parameters according to implicit user feedback (a single ``click'' on one of the images) until the user finds the target image he or she had in mind, thereby completing the retrieval process.

Although the mental image retrieval algorithm is a promising approach for query-free image retrieval, it has certain drawbacks in our application scenario, mainly because of the \textbf{variability in human decision-making}. Specifically, the clothing retrieval process is highly target-dependent and user-dependent. Usually, a garment is best described by certain attributes, such as a specially designed collar shape or a cartoon pattern, and garments are produced with great diversity in these attributes to impress customers. Moreover, even for the same image, different users at different times may form totally different interpretation, leading to high levels of subjectiveness and inconsistency. Therefore, existing methods, in which a fixed feature representation is used for images throughout the retrieval process, cannot effectively address all clothing retrieval scenarios.

%\begin{itemize}[leftmargin=*]
%\item {\bf Gaps between low-level image features and high-level user descriptions}. It is widely acknowledged that low-level image features are often inadequate for high-level image retrieval tasks. In the case of mental image retrieval, the choice made by the user is usually of high-level, which introduces the well-known ``semantic gap''.
%\item {\bf }. The clothing retrieval process is highly target-dependent and user-dependent. Usually a garment is best described by certain attributes, such as a special design on the collar shape or a cartoon pattern. Even for a same image, usually different users at different times may have diverse interpretation, showing high subjectiveness and inconsistency. Therefore, existing methods, which use a fixed feature representation of images throughout the retrieval process, could not address all the cases of clothing retrieval effectively.
%\end{itemize}

%While existing works mainly focus on retrieving items with explicit queries \cite{hadi2015buy,huang2015cross,vittayakorn2015runway},
%To account for the variability in human decision-making, w
We therefore propose a new mental image retrieval algorithm that simultaneously refines the retrieved images and estimates the users' criterion with respect to heterogeneous features extracted from multiple clothing attributes.
Unlike in previous studies \cite{fang2005experiments,ferecatu2007interactive,suditu2011heat,suditu2012iterative}, we model both the target image and the feature that drives the user's responses as random variables and update them in a unified Bayesian framework. The two random variables are not assumed to be conditionally independent in our algorithm. The processes of updating the posterior distributions are therefore highly intertwined and are performed simultaneously with the help of auxiliary conditional probabilities. For multiple clothing attributes (\textit{e.g.}, color, category, collar shape, skirt length), deep convolutional networks \cite{szegedy2015going} are adopted to train attribute-specific feature representations. By doing so, we maintain an independent similarity metric for each of the feature sets and iteratively update the weights of multiple features given the user's selections. As a result, for each individual search session, the system attempts to ``guess'' the metric that matches the user's answers, thus accounting for the variability in human decision-making.

%and present an algorithm that simultaneously refines the retrieved images and re-weights heterogeneous features

%At each iteration, the method selects a set of candidate images, shows them to the user, then updates the underlying model parameters according to the user feedback (a single ``click'' on one of the images) implicitly. An cumulative probability distribution is kept by the algorithm to account for the past interactions from the user, which gradually approaches the user's target through iterations until the target image is finally found in the selected candidate images.

%Although mental image retrieval algorithms have shown good potentials to solve query-free retrieval problems \cite{ferecatu2007interactive,suditu2011heat,suditu2012iterative}, they still have certain drawbacks for clothing retrieval, mostly according to the following two aspects:

The contributions of this paper are summarized as follows: 1) we investigate a new type of clothing retrieval task in which the target image resides only in the users' mind, which is a more realistic scenario for real-world applications; 2) by modeling this task as a mental image retrieval problem, we propose a new implicit relevance feedback algorithm that accounts for the variability of user selections through a novel Bayesian framework in which the target image and the ``user metric'' are treated as two random variables in a unified formulation; 3) we use deep convolutional features to represent multiple clothing attributes and analyze their impact in a case-sensitive manner according to user feedback; and 4) we study the effectiveness of the proposed algorithm by conducting comparative experiments on real users. The resulting weight distributions reveal a reasonable approximation of the ``user metric'' in realistic scenarios, in which high variability exists across individual search sessions.

The remainder of the paper is organized as follows. Section \ref{sec::relatedwork} presents related works. We introduce the update, answer and display models for our Bayesian framework in Section \ref{sec::algorithm}. Experimental results and detailed discussions are presented in Section \ref{sec::exp}. Finally, Section \ref{sec::conclusion} concludes the paper.

\section{Related Work}
\label{sec::relatedwork}

\textbf{Clothing retrieval.} A great deal of work has been done in the last few years on the subject of content-based image retrieval (CBIR)\cite{murala2012local,chu2013robust,chen2013scalable}, a typical domain of which is clothing recognition and recommendation \cite{liu2012hi,kalantidis2013getting,jagadeesh2014large}. In general, a clothing recognition system involves multiple steps, including human detection, clothing parsing, feature extraction, attribute prediction, and clothing recommendation. Given a query image of a person, state-of-the-art object detection \cite{girshick2014rich} or human pose estimation algorithms \cite{yang2011articulated} are usually adopted to extract foreground regions, followed by specifically designed clothing parsing methods \cite{yamaguchi2013paper,liu2014fashion,liang2015human,liang2016clothes,zhao2016clothing} to assign a semantic clothing label to each pixel in the image. Based on the detected regions at the bounding-box level or the pixel level, several works have used attribute-based frameworks to describe clothing items \cite{bossard2012apparel,chen2012describing,chen2015deep,abdulnabi2015multi}. The considered application scenarios are rather diverse: examples include predicting a user's social tribe \cite{kwak2013bikers,hong2015understanding}, occupation \cite{song2011predicting,shao2013you}, and aesthetics \cite{liu2012hi,kiapour2014hipster,simo2015neuroaesthetics,su2012preference}.

% Another related topic to our paper is the cross-scenario retrieval \cite{kalantidis2013getting,vittayakorn2015runway,huang2015cross} or specifically the ``street-to-shop'' problem \cite{liu2012street,hadi2015buy}, where the goal is to find an exact matching item from photos taken by customers on streets with complicated background and severe clutter to shop images taken by professional photographers with cleaner background and more regular poses. All of these methods need the users to provide an explicit street image as query; on the contrary, our method requires no query images and utilizes implicit relevance feedback to solve the retrieval problem.

\textbf{Relevance feedback.}
Relevance feedback (RF)
%%%Editor - An acronym or abbreviation is often defined at the first use of
%%%the related term in the main test and then used throughout the remainder
%%%of the manuscript. Please consider adhering to this convention.
was initially developed for use in document retrieval \cite{rocchio1966document} and was introduced into content-based image retrieval (CBIR) during the 1990s \cite{kurita1993learning}. Since that time, RF algorithms have been shown to enable drastic performance boosts in retrieval systems \cite{kherfi2006relevance,tao2006direct,auer2010pinview,lu2003joint,zhou2003relevance,grigorova2007content,broilo2010stochastic,bian2010biased,sun2010image,su2011efficient} and attribute learning \cite{kovashka2012whittlesearch, biswas2013simultaneous}.

In the context of feature weighting in RF, Rui \textit{et al.} \cite{rui1997content} proposed a re-weighting approach in which image feature vectors are converted into weighted-term vectors in MARS.
%%%Editor - Acronyms and abbreviations are often defined at their first use
%%%in the main text and then used throughout the remainder of the manuscript.
%%%Please consider adhering to this convention.
Another solution is to move the query point toward the contour of the user's preference in feature space, as is done, for example, in the famous Rocchio algorithm \cite{rocchio1966document}. The FA-RF method \cite{grigorova2007content} uses two iterative techniques to exploit relevance information: query refinement and feature re-weighting.
Recently, Jiang \textit{et al.} \cite{jiang2014zero} proposed a weighting scheme based on multiple modalities for zero-example video retrieval, in which logistic regression is applied given binary feedback.

Although feature re-weighting schemes are widely used in the literature on RF, most of these schemes are based on binary feedback, in which the user is asked to label positive and negative examples.
The proposed re-weighting scheme differs from previous works in terms of the problem formulation and the form of user feedback. We adopt a unified Bayesian RF framework and require minimum user feedback - only one click in each iteration. Therefore, a direct comparison with previous feature re-weighting approaches in RF is unfeasible.

\textbf{Mental image retrieval.}
Mental image retrieval, \textit{i.e.}, searching images without any explicit query, was pioneered by Cox \textit{et al.} \cite{cox2000bayesian}. They proposed a Bayesian framework on iterative relevance feedback to retrieve a specific image in the database (target search). Fang and German \cite{fang2005experiments} proposed an efficient display algorithm which only needs one click per iteration by the user, and applied it to mental face retrieval. Afterwards, Ferecatu \cite{ferecatu2007interactive} extended the framework to category search instead of target search. The application was scaled to large-scale datasets by Suditu and Fleuret \cite{suditu2011heat,suditu2012iterative} who adopted a hierarchical and expandable adaptive trace algorithm benefited from adaptive exploration/exploitation trade-off.
Similar to the idea of mental image retrieval, Auer \textit{et al.} \cite{auer2011relevance} maintained the weights of images by giving less relevant images a constant discount at each iteration.

Query-free methods show great potential in image retrieval applications. However, neither of them considers the variability of multiple features on different target images and different users. The proposed method can be regarded as a new exploration of introducing the idea of feature re-weighting into mental image retrieval and applying the algorithm into a more practical task of the clothing retrieval problem.

\section{Bayesian Retrieval Framework}
\label{sec::algorithm}
The core of the proposed algorithm is a feature re-weighting and target-modeling scheme in a Bayesian framework. In addition to modeling the target image as a random variable as done in previous studies \cite{fang2005experiments,ferecatu2007interactive,suditu2012iterative}, we treat the feature that drives the user's responses as another random variable and jointly refine the retrieved images and re-weight heterogeneous features in a unified Bayesian formulation. Our re-weighting scheme constructs a short-term approximation of the user metric, which is independently learned within each individual search session, to capture the inconsistent nature of user behavior.

\subsection{Preliminaries}\label{subsec::preliminary}
Suppose that there are $N$ images in a database $S$, say $I_1,...,I_N$. The target is a particular image in the database, denoted by $Y$. In the stochastic formulation, we define $Y$ as a random variable with some initial distribution $p_0(k)=P(Y=k),k\in S$. For this study, we empirically initialize this variable with a uniform distribution for equal probability, which means that each image has an equal probability of being the target image.

A relevance feedback session is composed of several iterations, each of which involves two quantities: a subset $D\subset S$ of $n$ displayed images, and the response of the user, denoted by $X_D\in D$. We assume that if the target image is displayed, the user will recognize it and terminate the search. Otherwise, when the target $Y$ is not in $D$, the user selects $X_D$ as the image that is closest to the target according to the metric in the user's mind.

Suppose there are $M$ sets of features, denoted by $F=1,2,...,M$. For each feature $j$, we compute a local similarity metric $s_j(x,y)$, where $x$ and $y$ are any two of the images in the database. To model the ``user metric'' in individual search sessions, we introduce another random variable $W$, which represents the probability of the user making decisions according to different features. $W$ is also given a uniform initial distribution $w_0(j)=P(W=j),j\in F$, representing that each feature has an equal probability of being the dominant one.

Let the user feedback up through the $t$-th iteration be denoted by
\begin{equation}
B_t=\bigcap_{i=1}^t\{X_{D_i}=x_i\},
\end{equation}
where $t=1,2,...$ is an index representing the iterations and $x_i$ is the user's selection in the $t$-th iteration.

The cumulative posterior distribution captures the effect of the user feedback accumulated through all previous iterations:
\begin{equation}
p_t(k)=P(Y=k|B_t),k\in S,
\end{equation}
where $p_t(k)$ represents the probability that image $k$ is the target image given the user feedback up through the previous $t$ iterations. A larger $p_t(k)$ indicates that the $k$-th image is more likely to be the target image and thus makes it more likely to be selected as one of the images to be displayed in upcoming sessions.

Analogously, we define
\begin{equation}
w_t(j)=P(W=j|B_t),j\in F,
\end{equation}
which represents the probability of the $j$-th feature being the dominant feature driving the user's responses throughout the previous $t$ iterations.

A critical assumption in our algorithm is that the random variables $Y$ and $W$ are not conditionally independent. Such an assumption is reasonable because the characteristics of the target image will naturally affect the way people remember that image. Therefore, we further define two auxiliary probabilities as follows:
\begin{equation}
\rho_t(k,j)=P(Y=k|B_t,W=j),k=1,...,N,
\end{equation}
\begin{equation}
\omega_t(j,k)=P(W=j|B_t,Y=k),j=1,...,M.
\end{equation}

Here, $\rho_t(k,j)$ represents the probability of image $k$ being the target after $t$ iterations if only feature $j$ is considered. $\omega_t(j,k)$ represents the probability of the user making his or her decisions based solely on the $j$-th feature after $t$ iterations conditioned on the target being $k$.
These auxiliary probabilities allow the cumulative posterior distributions to be decomposed into simpler representations that can be directly updated based on user feedback. Subsequently, the posterior distributions are updated based on the auxiliary probabilities.

Given the initial distributions $p_0(k)$ and $w_0(j)$, the proposed framework iteratively executes the following three basic steps:

\begin{itemize}
\item Update Model: updates the posterior distributions $p_{t+1}(k)$ and $w_{t+1}(j)$ based on the user's response in iteration $t+1$ and the previous evidence $p_t(k)$ and $w_t(j)$;
\item Display Model: determines which images to display in iteration $t+1$ based on the posterior distribution $\{p_{t+1}(k)\}$;
\item Answer Model: specifies the probability of the user selecting image $X_D\in D$ according to the metrics considered in the algorithm.
\end{itemize}

\subsection{Update Model}\label{subsec::updatemodel}
The update model is designed to update the posterior distributions of the target image and user metric given the user's feedback in iteration $t$. It consists of two phases: updating the auxiliary probabilities $\rho_t(k,j)$ and $\omega_t(j,k)$ and obtaining the posterior distributions $p_t(k)$ and $w_t(j)$ based on the new auxiliary probabilities.

Specifically, supposing that the user's response in one iteration, $X_{D_{t+1}}$, is conditionally independent of the previous feedback $B_t$ given the target being $k$ and a selected feature $j$, we update the auxiliary probabilities as follows:
\begin{eqnarray}\label{eqn::rho}
&&\rho_{t+1}(k,j)=P(Y=k|B_{t+1},W=j)\nonumber\\
&&=P(Y=k|B_t,X_{D_{t+1}}=x_{t+1},W=j)\nonumber\\
&&=\frac{P(Y=k,X_{D_{t+1}}=x_{t+1}|B_t,W=j)}{P(X_{D_{t+1}}=x_{t+1})}\nonumber\\
&&\propto P(Y=k|W=j,B_t)\cdot \nonumber\\
&& \quad P(X_{D_{t+1}}=x_{t+1}|Y=k,W=j)\nonumber\\
&&=\rho_t(k,j)P(X_{D_{t+1}}=x_{t+1}|Y=k,W=j).
\end{eqnarray}

Analogously,
\begin{eqnarray}\label{eqn::omega}
&&\omega_{t+1}(j,k) = P(W=j|B_{t+1},Y=k)\nonumber\\
&&=P(W=j|B_t,Y=k,X_{D_{t+1}}=x_{t+1})\nonumber\\
&&\propto \omega_t(j,k)P(X_{D_{t+1}}=x_{t+1}|Y=k,W=j).
\end{eqnarray}

Eq. \eqref{eqn::rho} and Eq. \eqref{eqn::omega} indicate that the process of updating the auxiliary probabilities can be regarded as taking the product of the previous auxiliary probabilities with the new evidence from the user response, $P(X_{D_{t+1}}=x_{t+1}|Y=k,W=j)$, followed by a normalization process.

Note that the posterior probabilities can be computed as linear combinations of the auxiliary probabilities using the law of total probability:
\begin{eqnarray}\label{eqn::p}
%p_t(k)&=&P(Y=k|B_t) \nonumber\\
p_t(k)&=&\sum_{j=1}^M P(Y=k|B_t,W=j)P(W=j|B_t) \nonumber\\
&=&\sum_{j=1}^M \rho_t(k,j)w_t(j),
\end{eqnarray}\vspace{-2ex}
\begin{eqnarray}\label{eqn::w}
w_t(j)&=&\sum_{k=1}^N P(W=j|B_t,Y=k)P(Y=k|B_t) \nonumber\\
&=&\sum_{k=1}^N \omega_t(j,k)p_t(k).
\end{eqnarray}

Let $\textbf{P}=[\rho_t]_{k,j}$ and $\textbf{W}=[\omega_t ]_{j,k}$, where $\textbf{P}$ and $\textbf{W}$ are matrices with dimensions of $N\times M$ and $M\times N$, respectively. Here, we omit iteration $t$ for brevity. Let $\vec{p}=[p]_{k}$ and $\vec{w}=[w]_j$, where $\vec{p}$ has dimensions of $N\times 1$ and $\vec{w}$ has dimensions of $M\times 1$.
Rewrite Eq. \eqref{eqn::p} and Eq. \eqref{eqn::w} as follows:
\begin{equation}\label{eqn::pn}
\vec{p} = \textbf{P} \vec{w},
\end{equation}
\begin{equation}\label{eqn::wn}
\vec{w} = \textbf{W} \vec{p}.
\end{equation}

From Eq. \eqref{eqn::pn} and Eq. \eqref{eqn::wn}, we obtain
\begin{equation}\label{eqn::solvep}
(\textbf{P} \textbf{W} - \textbf{I}_N) \vec{p}=\vec{0}_N,
\end{equation}
\begin{equation}\label{eqn::solvew}
(\textbf{W} \textbf{P} - \textbf{I}_M) \vec{w}=\vec{0}_M.
\end{equation}

Let $\textbf{G}=\textbf{P} \textbf{W} - \textbf{I}_N$ and $\textbf{H}=\textbf{W} \textbf{P} - \textbf{I}_M$. Then, $\vec{p}$ and $\vec{w}$ can be computed as the bases of the null spaces of $\textbf{G}$ and $\textbf{H}$, respectively.

\begin{algorithm}[ht]
\caption{Bayesian relevance feedback framework}
\label{alg::champ}
\begin{algorithmic} \renewcommand{\algorithmicrequire}{\textbf{Input:}} \renewcommand{\algorithmicensure}{\textbf{Output:}}
\REQUIRE
Target image $T$, randomly chosen initially displayed images $D_0$.
\ENSURE Number of iterations $t$ required to retrieve the target image.
\STATE Initialize the posterior distributions $\{p_0(k)\}$ and $\{w_0(j)\}$ as well as the auxiliary probabilities $\textbf{P}_0$ and $\textbf{W}_0$ with uniform distributions.
\REPEAT
\STATE (User feedback)
\STATE The user selects the image $x_t\in D_t$ that he or she thinks is the closest to the target image.
\STATE (Answer Model)
\STATE Compute the conditional probability of the user selecting $x_t$ using Eq. \eqref{eqn::answer}.
\STATE (Update Model)
\STATE Update the auxiliary probabilities using Eq. \eqref{eqn::rho} and Eq. \eqref{eqn::omega};
\STATE Update the posterior distributions using Eq. \eqref{eqn::solvep} and Eq. \eqref{eqn::solvew}.
\STATE (Display Model)
\STATE Select the $n$ images $D_{t+1}$ to be displayed by solving for the Voronoi partitions.
\STATE Go to the next round, $t=t+1$.
\UNTIL {The target image is included among the displayed images, $T\in X_D$}.
\end{algorithmic}
\end{algorithm}

\subsection{Answer Model}\label{answermodel}
Given $n$ displayed images, the answer model computes the probability of the user selecting each one of them. This probability is conditioned on two variables: the target and the feature used to measure the similarity. Intuitively, an image that is more similar to the target should be associated with a higher probability.

There are many possible strategies for requesting effective feedback from users. The most precise but also the most complicated strategy is to ask the user to provide a detailed value that measures the degree of similarity between each displayed image and the target. Alternatively, users can rank the displayed images according to their similarity to the target or can simply label each image as ``relevant'' or ``not relevant''. Following \cite{fang2005experiments}, we employ a scheme that involves minimum user interference called ``one-click'' feedback, in which the user only needs to select the image from among the $n$ displayed images that is ``closest'' to the target image according to the user's own ``metric''. Suppose that the target is not included among the displayed images. Let $s_j(x,j)$ be the similarity between two images $x$ and $y$ with respect to the $j$-th feature. For $i\in \{1,...,n\}$, we assume that the probability of the user selecting image $i$ as his or her response is
\begin{equation}\label{eqn::answer}
P(X_D=i|Y=k,W=j)=\frac{s_j(i,k)}{\sum_{l\in D}s_j(l,k)}.
\end{equation}

Consequently, conditioned on image $k$ being the target and based on feature $j$, the closer an image $i$ is to $k$, the more likely the user is to select it. In other words, if the user selects an image that has a low probability as computed by the answer model, the reason may be either that the $k$-th image is not the target or that the $j$-th feature is not the dominant feature considered by the user.

\subsection{Display Model}\label{subsec::displaymodel}
The display model determines the images to be displayed in the $t$-th iteration according to the posterior distribution $\{p_t(k)\}$. Whereas the update model and the answer model are used to determine which images are more important to the search session, the display model addresses the trade-off between exploration and exploitation in retrieving the target image. The most straightforward scheme is to select the $n$ images with the highest $p_t$. However, this scheme is not optimal since the images with the highest probabilities may be largely identical to each other. Instead, we adopt a strategy based on entropy, which minimizes the uncertainty of the target given the search history $B_t$ and the new answer $X_{D_{t+1}}$:

\begin{eqnarray}
D_{t+1}&=&\argmin_{D\subset S}H(Y|B_t,X_D)\nonumber\\
&=&\argmax_{D\subset S}I(Y;X_D|B_t).
\end{eqnarray}

Since $Y$ determines $X_{D_{t+1}}$,
\begin{equation}
D_{t+1}=\argmax_{D\subset S}H(X_D|B_t).
\end{equation}

The maximum entropy is achieved with a uniform distribution. Heuristically, we wish to choose $n$ seed images, say $\{x_1,...,x_n\}$, such that the partitions driven by computing the nearest seed, which can be modeled as Voronoi partitions, have nearly equal mass under the posterior distribution. In our framework, users select images based on different features under a distribution denoted by $\omega_t(j,k)$. Therefore, the Voronoi partitions cannot be computed directly using a unified similarity metric. Hence, we use the expectation of the similarity under the posterior distribution as the metric.

Given two images $x$ and $y$, the expectation of the similarity is
\begin{eqnarray}
%s(x,y)&=&E(P(W=j|B_t,Y=x)\cdot P(W=j|B_t,Y=y)\cdot s_j(x,y))\nonumber\\
s(x,y)&=&\frac{\sum_{j=1}^M \omega_t(j,x)\omega_t(j,y)s_j(x,y)}{\sum_{j=1}^M \omega_t(j,x)\omega_t(j,y)}.
\end{eqnarray}

We compute a heuristic solution using a process identical to that in \cite{fang2005experiments} except that we use the expectation of the similarity. We omit the details of the implementation here because of space limitations. Algorithm \ref{alg::champ} summarizes the proposed Bayesian framework.

\section{Experiments}\label{sec::exp}
In this paper, we focus on the target search problem with respect to mental images, for which the goal is to retrieve a specific target that resides only in the user's mind via implicit relevance feedback. In addition to the exact target retrieval problem, the feature re-weighting scheme adopted in our method can also be employed for the retrieval of similar images from a larger dataset using the methods described in \cite{ferecatu2007interactive,suditu2011heat}; this task will be the focus of our future work. We conducted a user study to evaluate the effectiveness of the proposed re-weighting scheme; here, we present discussions of the experimental results.

\subsection{Dataset and Features}\label{subsec::dataset}
We collected a large-scale clothing item dataset with detailed attribute-level annotation for use in our experiments. Specifically, we crawled images from several large online shopping websites, including TMALL.com and TAOBAO.com, resulting in a total of approximately 300,000 images. Each image is associated with a set of attributes provided by the websites, ranging from generic properties such as garment categories and colors to specific attributes such as skirt length and button shape.

\begin{figure}[t]
\centering
\includegraphics[width=.45\textwidth]{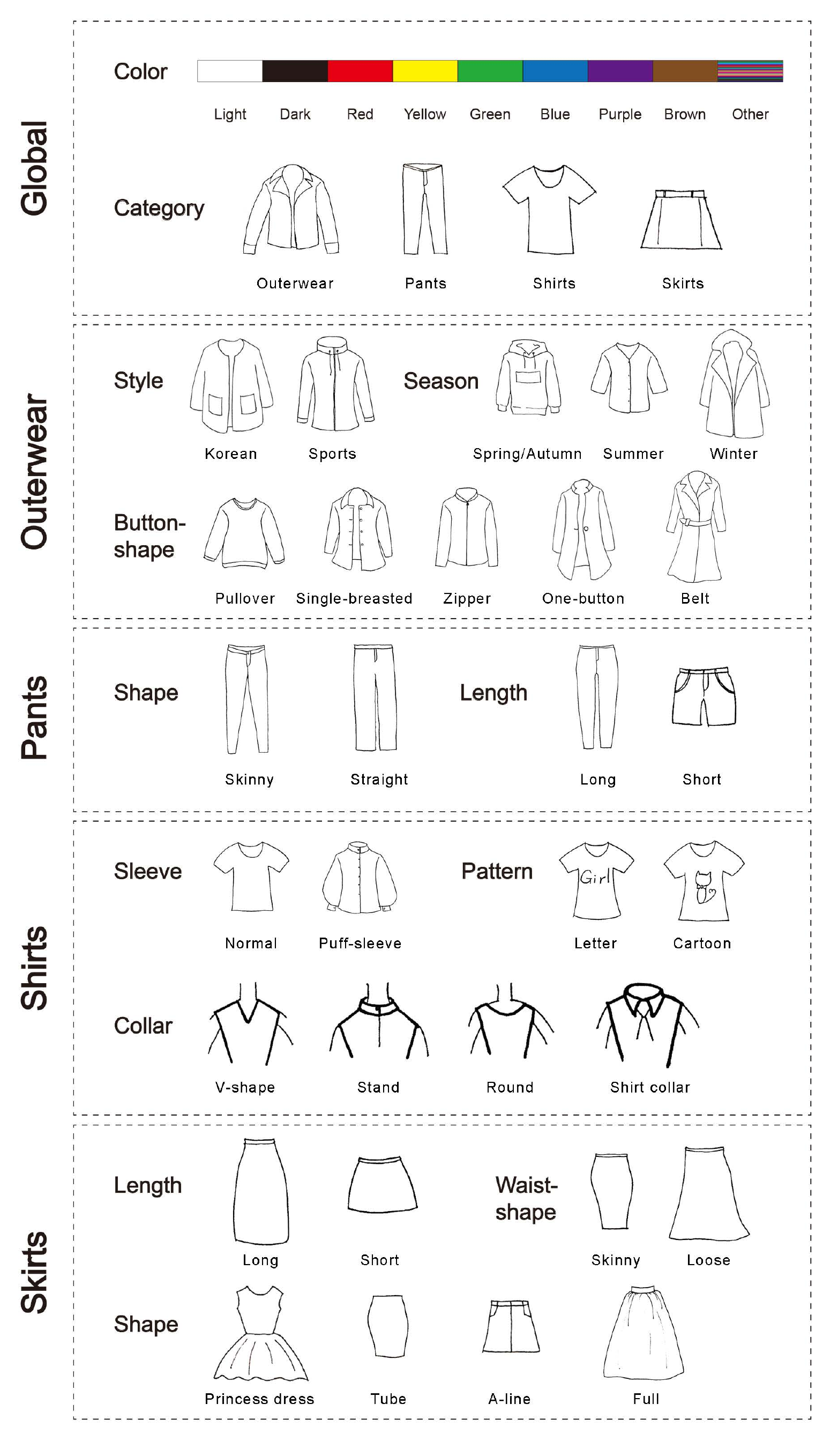}
\caption{Illustration of the dataset and the clothing attributes used in our experiments.}
\label{fig::dataset}
\end{figure}

In our real-user experiments, the users were requested to provide a rough selection of the garment category\footnote{Four entries: shirts, outerwear, pants, and skirts.} and color\footnote{Nine entries: light, dark, red, yellow, green, blue, purple, brown, and other.} of their target items. We regard this request as a simple annotation that a motivated user could easily provide. After this rough filtering step, most subsets in our dataset contained 500-1500 images. For each category, we selected the five most relevant attributes for the proposed feature re-weighting algorithm. An illustration of the dataset and the attributes used in our experiments can be found in Figure \ref{fig::dataset}.

Motivated by recent achievements attained by using deep convolutional networks (CNNs) for various image recognition tasks \cite{girshick2014rich,razavian2014cnn}, especially for attribute prediction \cite{zhang2014panda,escorcia2015relationship,abdulnabi2015multi,shankar2015deep,huang2015part}, we extracted domain-specific features using CNNs fine-tuned on our clothing dataset. Specifically, starting from the GoogleNet \cite{szegedy2015going} model pre-trained on ImageNet \cite{deng2009imagenet}, we performed two stages of fine-tuning. The first stage was conducted at the product level; \textit{i.e.}, the objective was to classify garment objects into different products, where each product corresponds to several items of different colors or images taken from multiple shooting angles and at different scales. This stage is similar to the corresponding stage of the DeepID2 \cite{sun2014deep} architecture for face recognition, with the goal of extracting generic features related to the garment analysis task. In the subsequent stage, we further fine-tuned the CNN features on different attributes starting from the model obtained in the first stage. These features were then employed in our re-weighting algorithm to provide specific knowledge regarding multiple clothing attributes.

\begin{figure*}[!t]
\centering
\includegraphics[width=.95\textwidth]{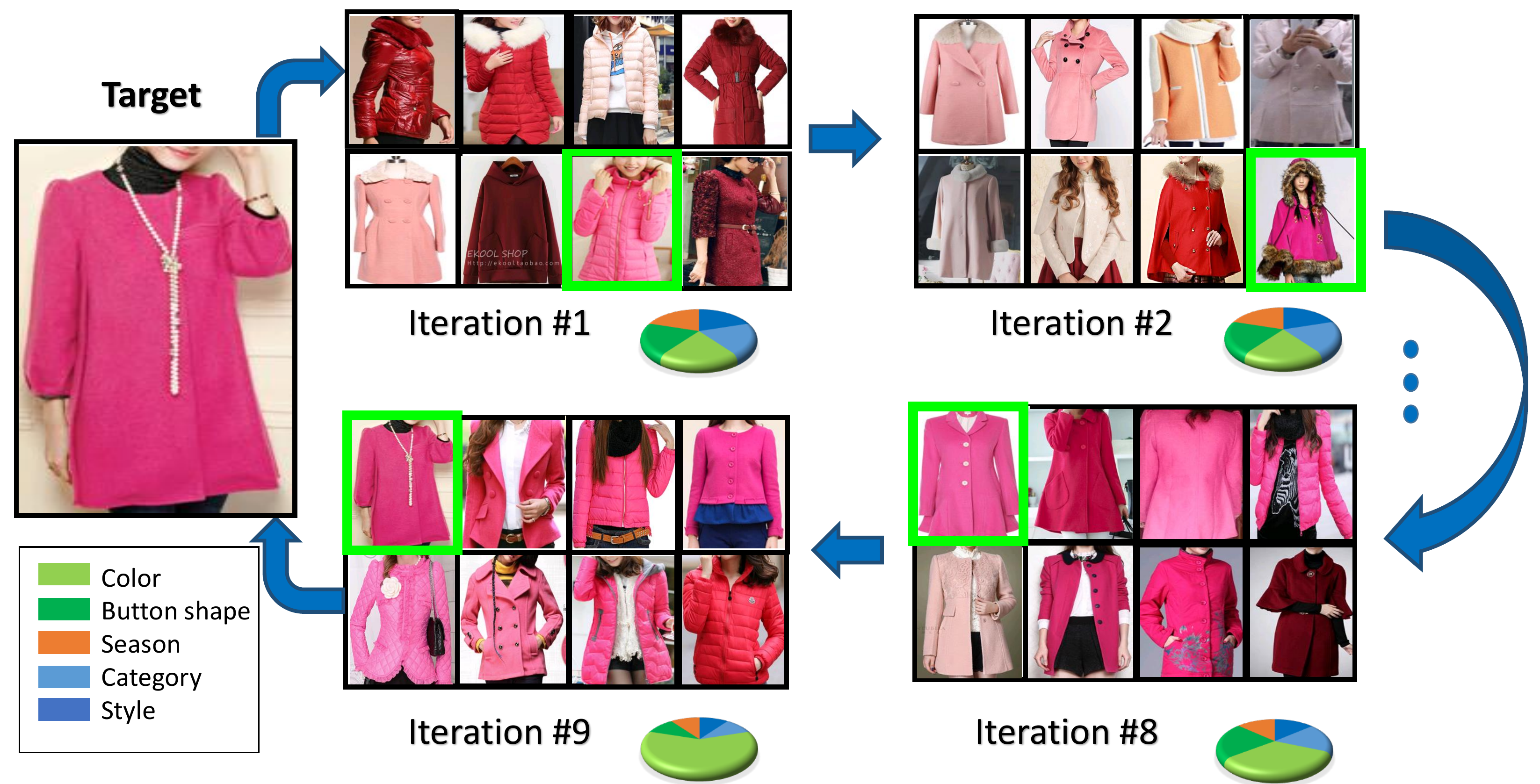}
\caption{Typical example of the relevance feedback procedure as conducted in the user study and the resulting weight distributions. In iteration 1, we display initial images at random and assign equal weights to all features. In the following iterations, the user selects the image that he or she thinks is closest to the target, as indicated by the highlighting of an image with a green box. The cumulative weight distributions learned by the algorithm up through the current iteration are illustrated by the pie charts below the candidate images. }
%The final weight distributions reveal that the first target image is retrieved mainly according to color (snow white), while the second image relies more on semantic information (plant and water). Better viewed in color.
\label{fig::demo}
\end{figure*}

\subsection{Evaluation Criteria}
In our user studies, we simulated the exact retrieval process for mental images by presenting a random target image to the user to use as a query while not providing the algorithm with any information about that query image. We refer to a single retrieval session with a specific target image as one ``mental image game''. For each game, there are three possible ending statuses:

\begin{itemize}
\item \textbf{Approved by System (AS)}: This status means that after several iterations, the exact target image is selected as one of the candidate images presented to the user, thereby completing the retrieval process.
\item \textbf{Approved by User (AU)}: Often, there are several images in the dataset that are highly similar to each other. Therefore, we added a ``find it'' button to our system, which allows the user to indicate that the target image has indeed been found even if the exact target image has not been selected for display.
\item \textbf{Game Abandoned (GA)}: This status accounts for the case in which a user becomes frustrated with the game and does not want to continue anymore. The algorithm is considered to have failed in this case.
\end{itemize}

For all retrieval sessions, we are concerned with the success rate (the ratio of the number of games that end as AS or AU to the total number of games); for all successful games, the key evaluation criterion is the number of relevance feedback iterations required to retrieve the target image, denoted by $T$. Following previous studies \cite{fang2005experiments,ferecatu2007interactive}, we evaluate the average number of iterations $E(T)$ and the cumulative probability $P(t\leq T)$. Better performance is indicated by a smaller value of $E(T)$ and a faster rate of increase of $P(t\leq T)$.

%Meanwhile, to compare the proposed re-weighting scheme with the baseline using fixed feature representations \cite{fang2005experiments}, we conducted pairwise one-on-one comparison experiments using the same target image and initialized candidate images. As a result, we are able to evaluate the two schemes according to the winning rate, abandonment rate and averaging time needed for a successful retrieval session. They are also provided in our experiments as supporting metrics.

\subsection{Ideal User and Random User}\label{subsec::ideal}
There are two extreme cases of user decision-making: the ``ideal user'' and the ``random user''. The random user selects images at random. On the opposite end of the spectrum, the ideal user always selects an image in strict accordance with the ``system metric''; \textit{i.e.}, given a target $q$, the ideal user selects $x=\argmax_{i\in D}P(X=i|Y=q)$ from among the displayed images, where
\begin{equation}
P(X_t=i|Y=q)=\sum_{j=1}^M \omega_t(j,q)\frac{s_j(i,q)}{\sum_{i\in D_t}s_j(l,q)}.
\end{equation}
In this paper, we assume that the ideal user can observe the underlying weights of different features in the system and consistently select the optimal image according to the system weight in consecutive iterations. In other words, the ideal user can perfectly monitor and make decisions following the ``system metric''. Therefore, the ideal user serves as an upper bound on the system performance.

\begin{figure*}[t]
\centering
\includegraphics[width=.9\textwidth]{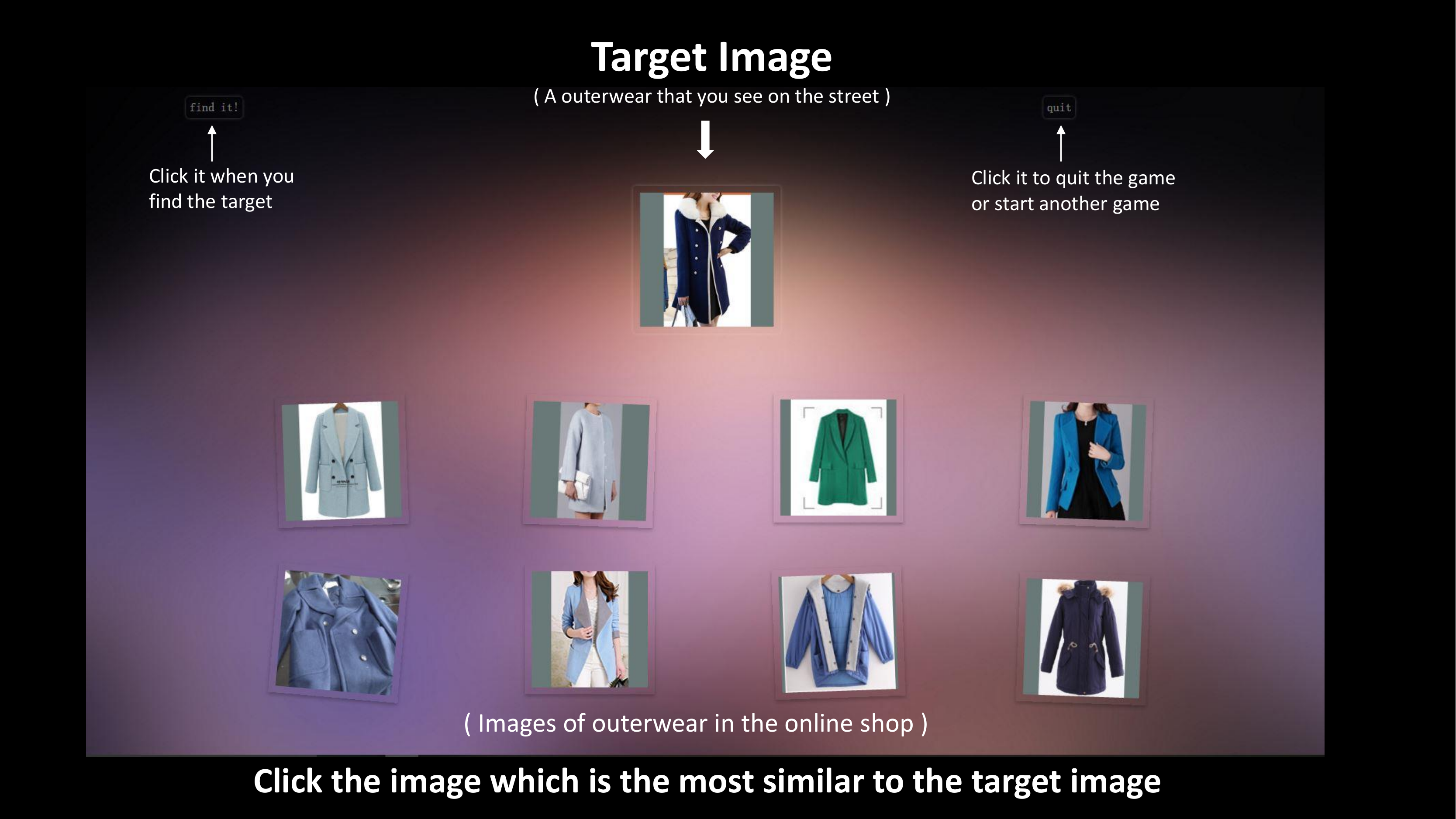}
\caption{User interface with instructions for our experiments, packaged as a ``mental image game''.}
\label{fig::gui}
\end{figure*}

\begin{figure*}[!t]
%\begin{minipage}{0.32\textwidth}
%\centering
%\includegraphics[width=.85\textwidth]{stats}
%\caption{Statistics of different ending status for experiments on real users. GA, AS, AU stands for game abandoned, approved by system, and approved by user.}
%\label{fig::stats}
%\end{minipage}
\begin{minipage}{0.32\textwidth}
\centering
\includegraphics[width=\textwidth]{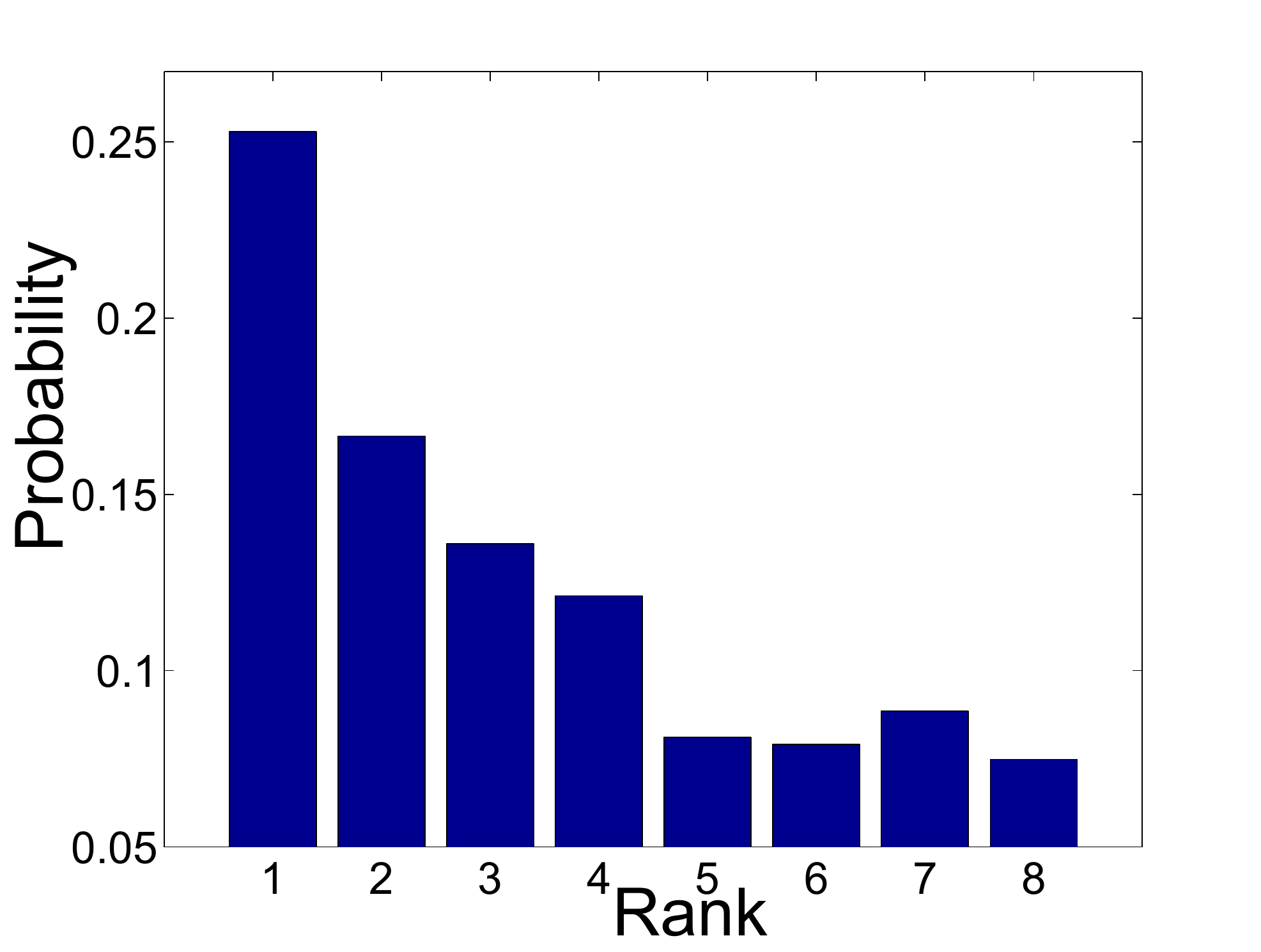}
\caption{The probability that the user will select the $m$-th closest image to the target according to the system metric.}
\label{fig::click}
\end{minipage}
\hfill
\begin{minipage}{0.32\textwidth}
\centering
\includegraphics[width=.95\textwidth]{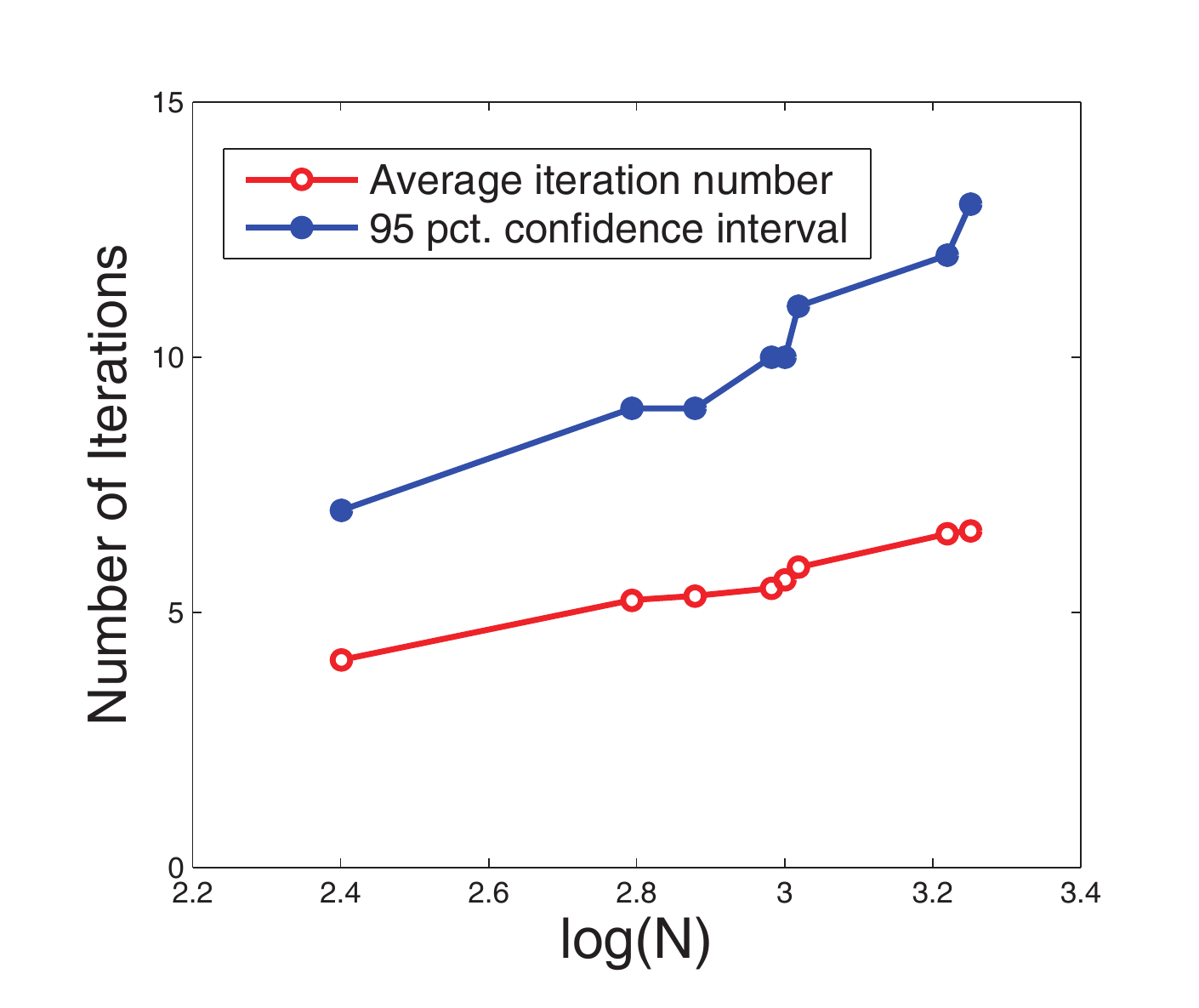}
\caption{Influence of the database scale $N$. The average number of search iterations is approximately log-linear with respect to $N$.}
\label{fig::scale}
\end{minipage}
\hfill
\begin{minipage}{0.32\textwidth}
\centering
\includegraphics[width=\textwidth]{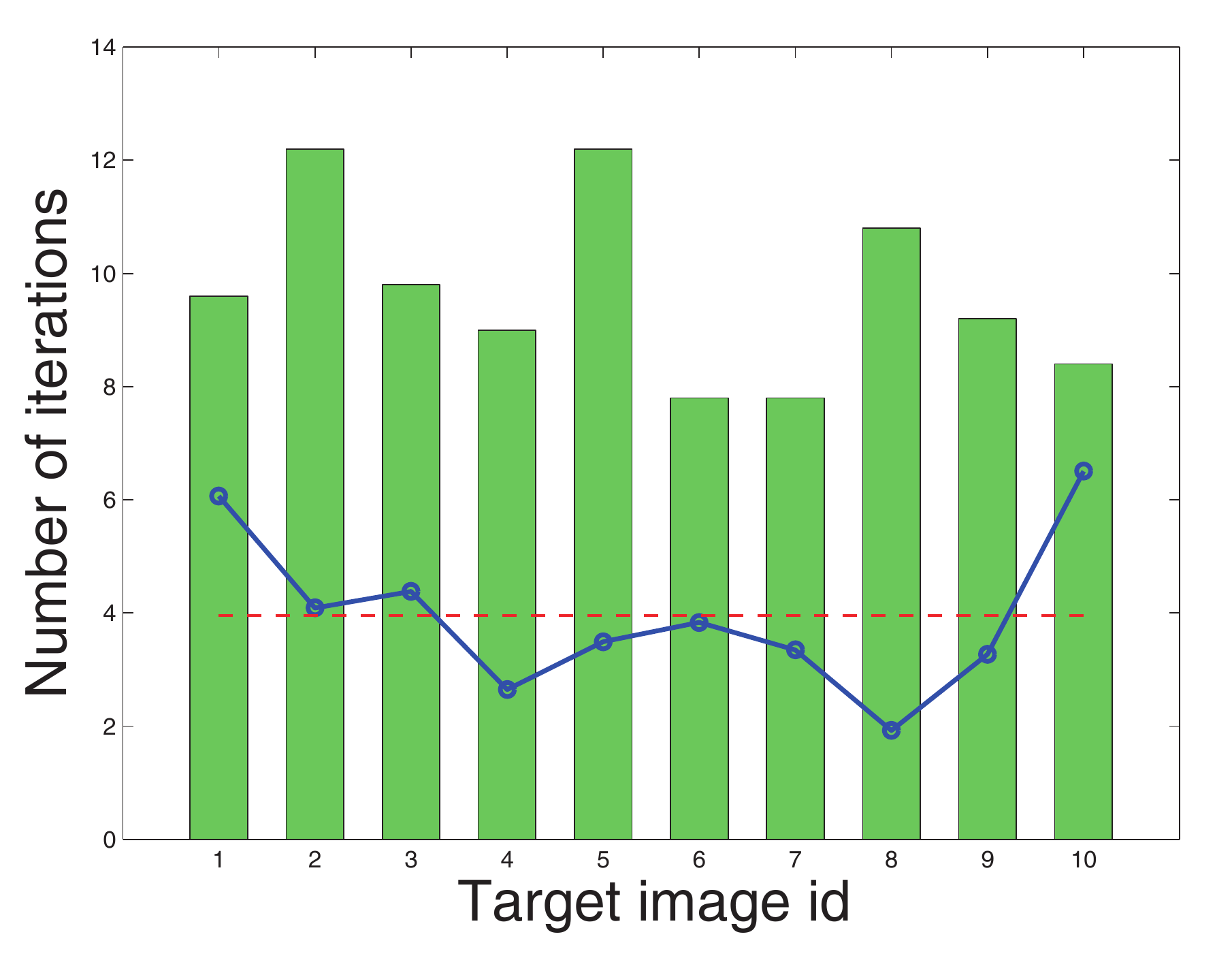}
\caption{The mean and standard deviation of the number of iterations when different initial images are displayed for the same target image.}
\label{fig::weight}
\end{minipage}
\end{figure*}

\subsection{Experiments with Real Users}\label{subsec::realuser}
To evaluate the performance of the proposed algorithm in realistic scenarios, we conducted a user study with 310 volunteers who were not familiar with the algorithm. To simulate the mental search scenario, for each search session, the user was first asked to provide generic attributes as inputs for filtering, and a target image was then shown on the screen to represent the image residing in the user's mind. The system begins by displaying a set of randomly selected images $D_0$, followed by iterations of user feedback and system updating until the target image is retrieved. We set $|D|=8$; \textit{i.e.}, eight images were displayed in each iteration.

To enable an in-depth analysis of the proposed algorithm, we conducted the experiment on four subsets of the clothing item dataset, corresponding to shirts, outerwear, pants and skirts, containing 9103, 8080, 9452 and 6130 images, respectively, and divided each category into nine colors. Each subcategory contained 500-1500 images, which is a typical size for a specialized online store.
Figure \ref{fig::demo} shows a typical search session and the resulting weight distributions obtained using the proposed algorithm. The algorithm reasonably simulates the predominant criterion that drives a user's responses. Notably, significant differences exist between the weight distributions generated in different cases. This finding reveals that the proposed algorithm assigns customized weights for each individual search session, thereby explicitly modeling the nature of human decision-making.
%%%Editor - Please note that it is unclear whether the term ``session'' is used
%%%here to refer to a complete game or to a single iteration within a game. From
%%%context, it seems that the intent might be to state that the weights are
%%%recalculated for each iteration; however, previously in the manuscript, it
%%%is explicitly stated that a ``session'' consists of several ``iterations''.
%%%Please be consistent with the usage of the term ``session'' throughout the
%%manuscript. For clarity, it might be preferable to eliminate the term
%%%``session'' and replace each instance of this term with either ``game'' or
%%%``iteration'', as appropriate.

The mechanism of the algorithm can be explained as follows. As an example, consider the ideal user who represents the system metric. Suppose that $j$ is the feature that has been assigned the highest weight after the first iteration. In subsequent iterations, the ideal user will tend to consistently select candidates according to $j$, which introduces a large bias in the first several rounds. As the search continues, this ``important'' feature becomes saturated; \textit{i.e.}, all images with a high posterior probability are similar in terms of this feature. Consequently, the user will up-weight other features with higher discriminative power. As a result, the algorithm has a ``zigzagging'' nature - the features repeatedly swing between under-saturation and over-saturation. Meanwhile, the cumulative probabilities capture the effect of all previous user interactions. Real users follow a similar pattern throughout the feedback process, although with more subjectiveness and inconsistency.

In addition to the performance of the re-weighting scheme proposed in this paper, we also present the results of a mental image retrieval algorithm \cite{fang2005experiments} that uses the generic clothing features obtained in the first stage of feature fine-tuning (see Section \ref{subsec::dataset}) and the results of later fusion based on multiple features for comparison.
As described previously, we perform this comparison based on a ``mental image game'' such as that shown in Figure \ref{fig::gui}. The comparative experiment was performed as follows: 1) Users were requested to complete games by clicking on the image that was most similar to their mental image. 2) If a displayed image was sufficiently similar to the target image according to the user, he or she could click the ``find it'' button to end the game immediately; otherwise, the game would continue until the exact target image appeared. 3) The user could abandon the game if it required too many iterations. We have provided an online demo to illustrate the experimental process\footnote{Demo link: 202.120.39.165:1333}.

We collected data from 926 retrieval sessions performed by real users for the comparative experiment: 312 sessions using the proposed algorithm, 325 using the algorithm presented in \cite{fang2005experiments}, and 289 using the later fusion scheme. Table \ref{tab::rate} shows the distributions of the game-ending statuses for the three algorithms. In general, the proposed algorithm achieved the lowest abandonment rate of $19.6\%$ compared with $27.7\%$ for the fixed-weight method and $25.3\%$ for the later fusion scheme. Moreover, the AU rate for the proposed algorithm was $18.9\%$, compared with $15.1\%$ for the fixed-weight method and $9.7\%$ for the scheme of late fusion, which indicates that the proposed algorithm can more closely approximate the high-level user metric. %Among all the recorded games, the proposed algorithm wins $60.5\%$ of the games, loses $33.1\%$, while the other $6.4\%$ are tied.
These findings show a rather consistent advantage of the proposed algorithm compared with the baseline method based on a fixed feature representation and the later fusion approach based on multiple features over all sessions and iterations. As benchmarks, the results for the “ideal user” and the “random user” under the same experimental conditions are also presented. Figure \ref{fig::click} depicts the rank distribution of the user selections according to the similarity metric maintained by the system. Although the correlation is not high, the distribution shows coherence between the users' selections and the system metric, meaning that sufficient information is conveyed to yield reasonable search results.

\begin{table}[t]
\caption{.}
\label{tab::rate}
\small
\centering
\vspace{+1ex}
\begin{tabular}{|l|c|c|c|c|}
\hline
  & GA & AS & AU & Success rate\\
\hline
Re-weight & 61 & 192 & 59 & $80.45\%$ \\
\hline
Fixed-weight & 90 & 186 & 49 & $72.31\%$ \\
\hline
Later fusion & 73 & 188 & 28 & $74.74\%$ \\
\hline
\end{tabular}
\end{table}

\begin{figure}[!t]
\centering
\includegraphics[width=.48\textwidth]{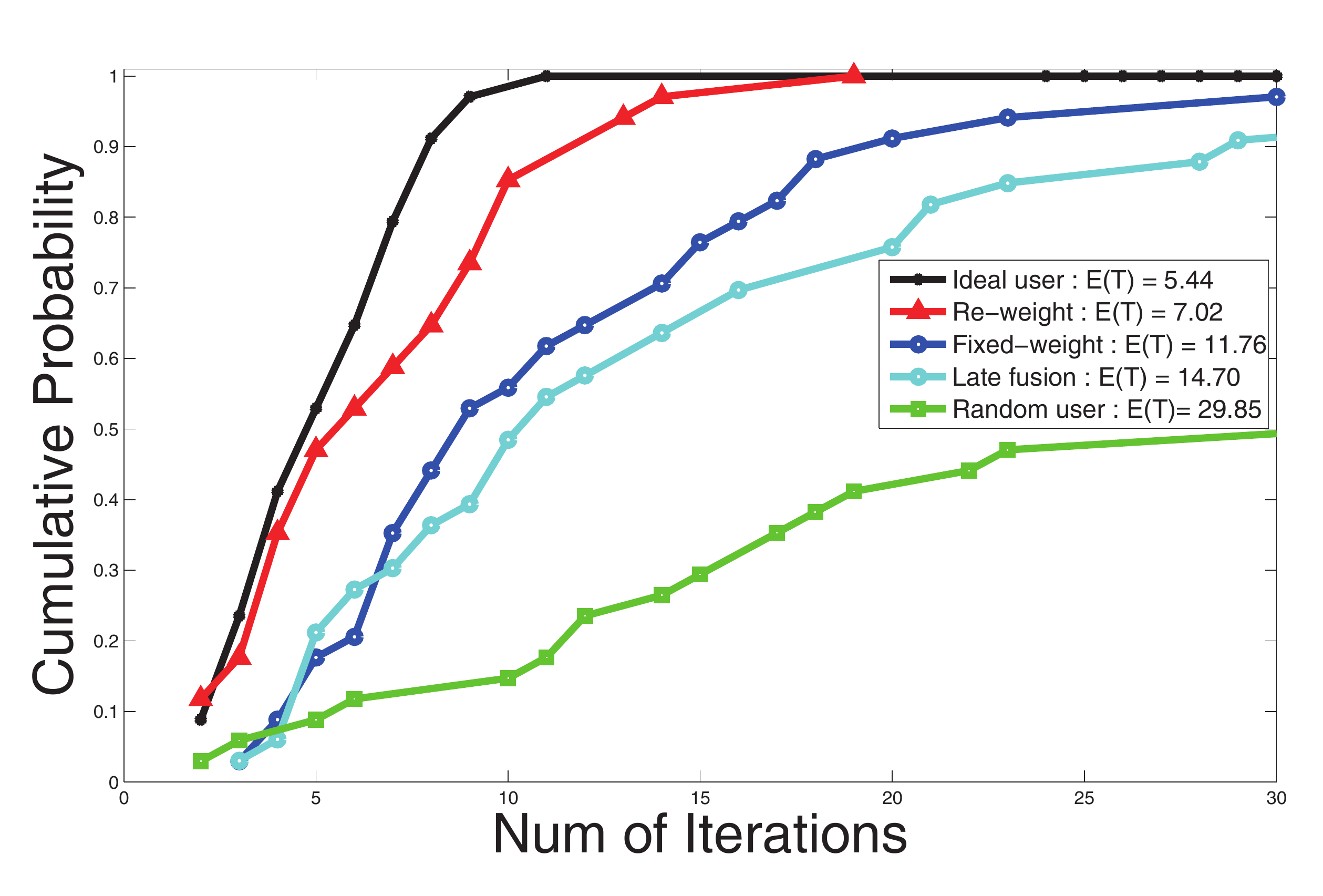}
\caption{The cumulative distributions of success probability with respect to search time for ideal, real and random users for the shirt subset of the database. The proposed algorithm is labeled as ``Re-weight''. We compare our results with those of \cite{fang2005experiments} and of the later fusion approach based on multiple features. }
\label{fig::flow}
\end{figure}

\begin{figure}[!t]
\centering
\includegraphics[width=.48\textwidth]{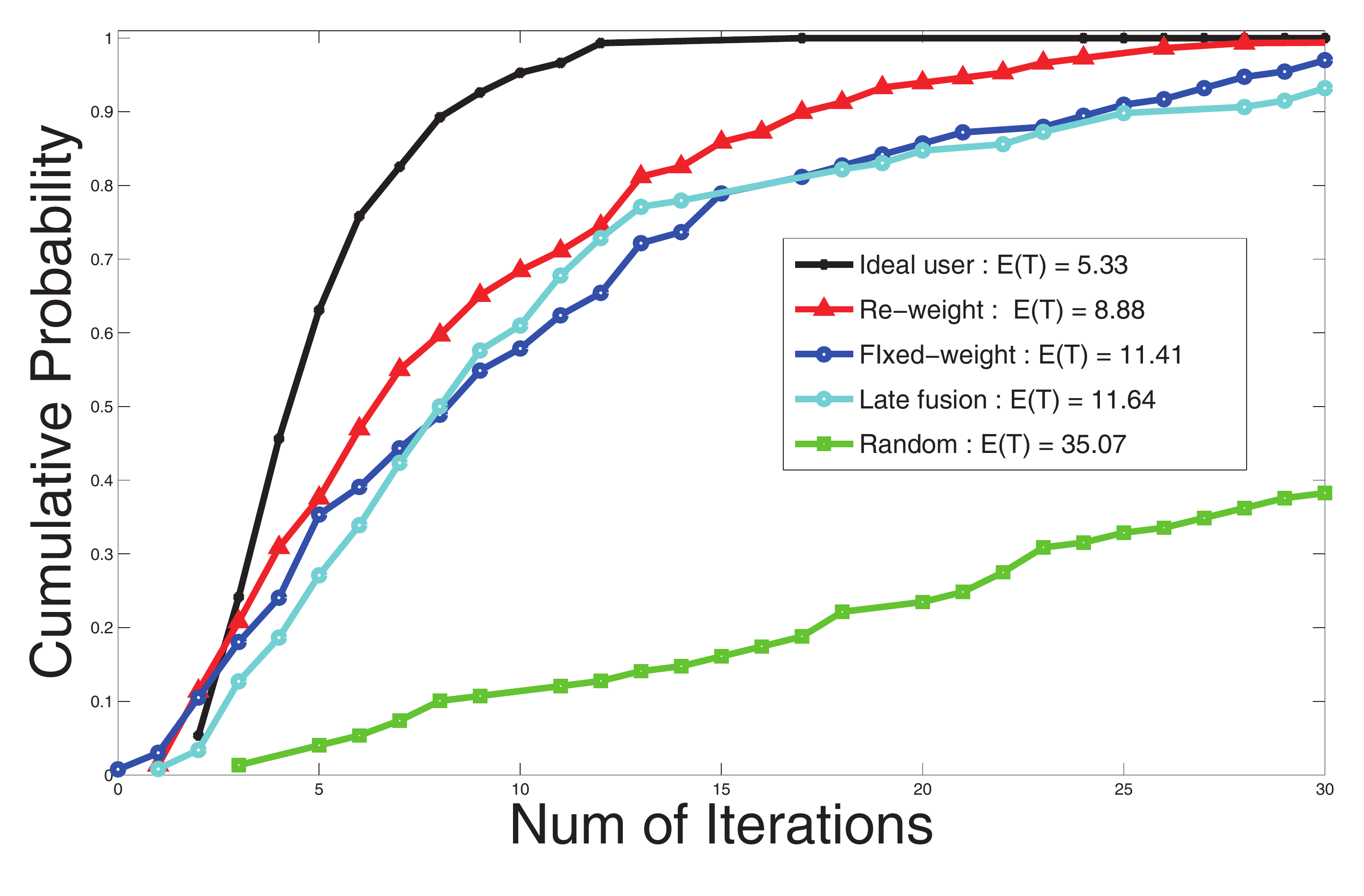}
\caption{The cumulative distributions of success probability with respect to search time for ideal, real and random users for the outerwear subset of the database. The proposed algorithm is labeled as ``Re-weight''. We compare our results with those of \cite{fang2005experiments} and of the later fusion approach based on multiple features. }
\label{fig::flow2}
\end{figure}

\begin{figure}[!t]
\centering
\includegraphics[width=.48\textwidth]{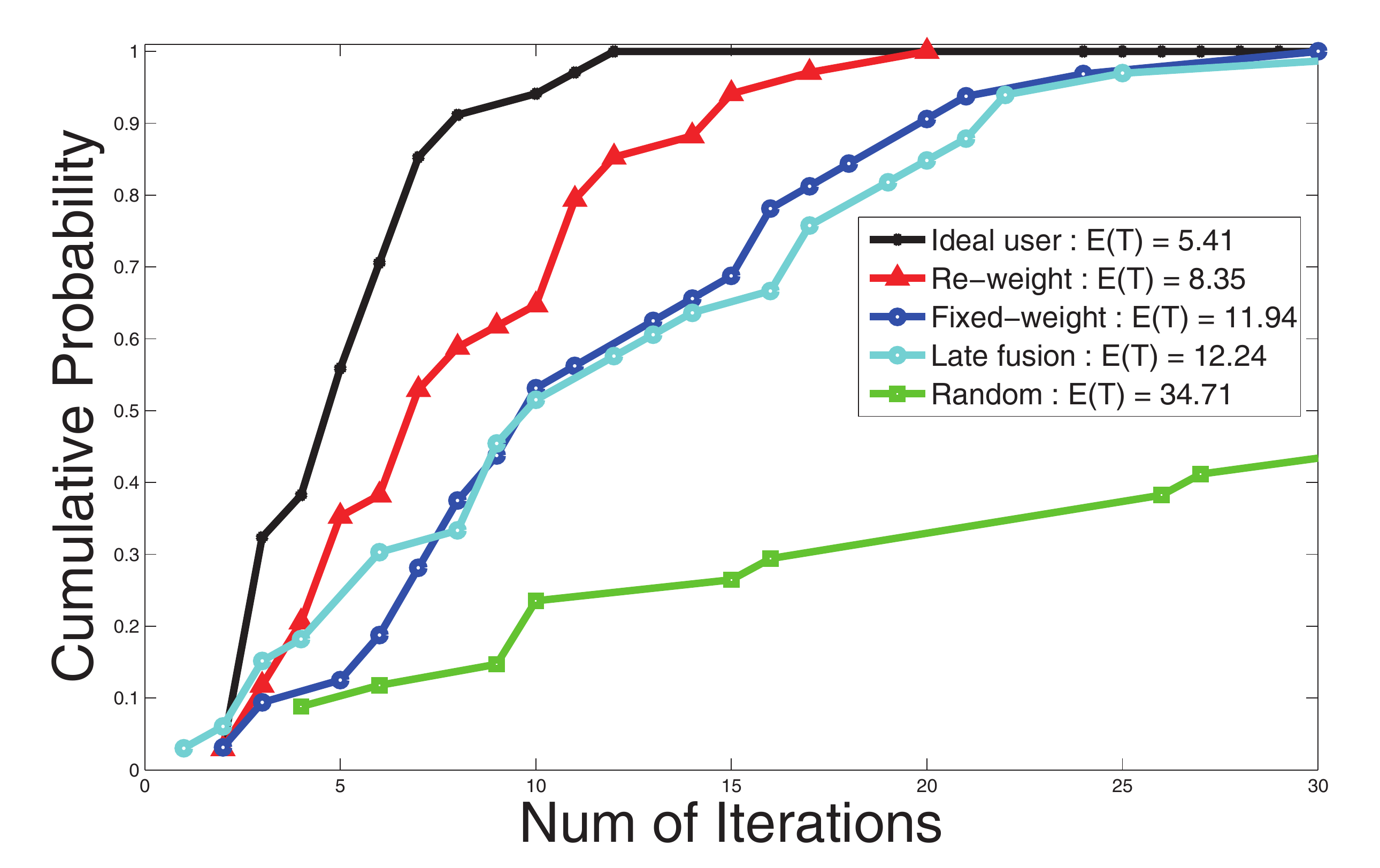}
\caption{The cumulative distributions of success probability with respect to search time for ideal, real and random users for the pants subset of the database. The proposed algorithm is labeled as ``Re-weight''. We compare our results with those of \cite{fang2005experiments} and of the later fusion approach based on multiple features. }
\label{fig::flow3}
\end{figure}

\begin{figure}[!t]
\centering
\includegraphics[width=.48\textwidth]{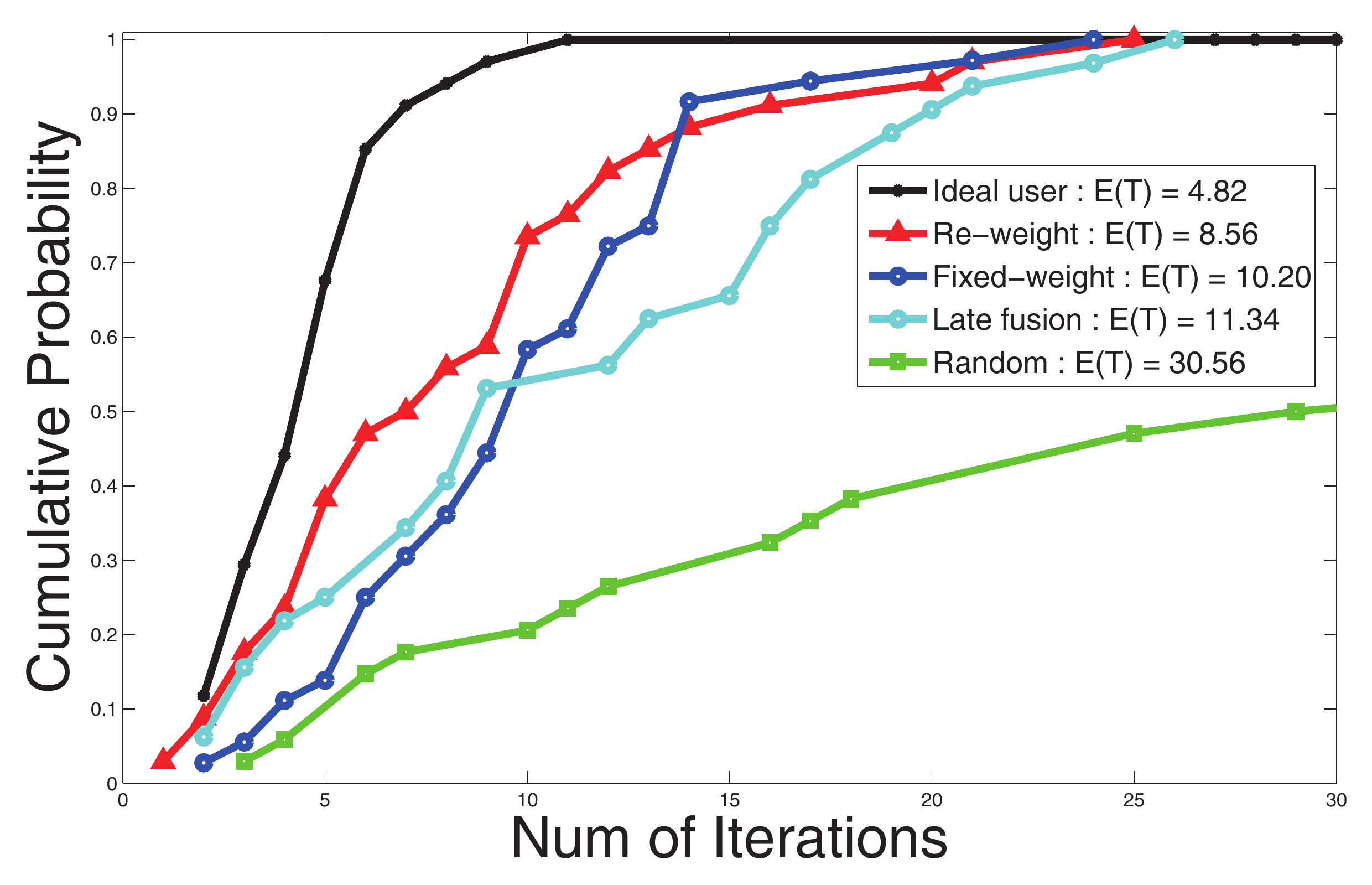}
\caption{The cumulative distributions of success probability with respect to search time for ideal, real and random users for the skirt subset of the database. The proposed algorithm is labeled as ``Re-weight''. We compare our results with those of \cite{fang2005experiments} and of the later fusion approach based on multiple features. }
\label{fig::flow4}
\end{figure}

Figure \ref{fig::flow}, Figure \ref{fig::flow2}, Figure \ref{fig::flow3} and Figure \ref{fig::flow4} show the cumulative probability distributions of the required number of iterations for the ideal user, the random user and real users on the four subsets of the clothing item dataset, namely, shirts, outerwear, pants and skirts. Obviously, relevance feedback contributes to improving the retrieval performance: real users need, on average, fewer than half of the iterations required by the random user. Compared with the method based on a fixed feature representation and the later fusion method, the proposed re-weighting scheme achieves better performance for real users on all four subsets, requiring only $7.02$, $8.88$, $8.35$ and $8.56$ iterations on average for shirts, outerwear, pants and skirts, respectively.

In particular, the proposed algorithm achieves markedly better performance for shirts and pants; thus, it can be presumed that when the subset of interest is larger, the proposed algorithm can achieve a higher performance improvement over the two baseline algorithms. By contrast, the subset of images representing skirts consists of only 6130 images. Detailed analysis shows that for the four categories on average, approximately $70\%$ of searches can be successfully completed within $10$ iterations; the percentage of successful completion increases to $95\%$ at $20$ iterations. Meanwhile, real users require, on average, one and a half times the number of iterations required by the ideal user to complete a search session. This discrepancy can be regarded as the gap between the ``system metric'' and the predicted ``user metric''.

%\caption{Statistics of different ending status for experiments on real users. GA, AS, AU stands for game abandoned, approved by system, and approved by user, respectively. For example, the entry of the second row and first column shows the number of games in which the user finds the target image using the proposed re-weighting algorithm, while fails to retrieve the target using the baseline method.}

\subsection{Extensive Analysis}
In addition to the performance evaluation and comparison with previously developed algorithms, here we present further discussions of various aspects of the proposed algorithm, including the average feature weight, the effect of the scale of the online image dataset, and the consistency among individual search sessions for a fixed target image.

\subsubsection{Average Feature Weight}
A significant property of the proposed algorithm is that it can automatically tune the weights of multiple features during a relevance feedback session. Although the resulting weights vary among individual sessions, the average feature weight provides insight into the discriminative power of different features.

\begin{table}[t]
\caption{Comparison of the average feature weights for real users and simulated ideal users using the proposed algorithm. The discriminative power of each individual feature is represented by the average number of search iterations required by the ideal user using each feature independently.}
\label{tab::weight}
\small
\centering
\vspace{+1ex}
\begin{tabular}{|l|c|c|c|c|c|}
\hline
  & Category & Style & Btn. Shape & Season & Color  \\
\hline
Real & 0.179 & 0.185 & 0.184 & 0.198 & 0.254 \\
\hline
Ideal & 0.219 & 0.212 & 0.213 & 0.176 & 0.180 \\
\hline
\hline
\textit{Avg. Iter.} &\textit{6.41} &\textit{5.59} &\textit{5.67} &\textit{6.66} &\textit{6.23} \\
\hline
%\multirow{5}{1.5cm}{Individual Feature Weight}
%   &    1     &    0    &     0    &     0     &     0     &  6.41 \\
%   \cline{2-7}
%   &    0     &    1    &     0    &     0     &     0     &  5.59 \\
%   \cline{2-7}
%   &    0     &    0    &     1    &     0     &     0     &  5.67 \\
%   \cline{2-7}
%   &    0     &    0    &     0    &     1     &     0     &  6.66 \\
%   \cline{2-7}
%   &    0     &    0    &     0    &     0     &     1     &  6.23 \\
% \hline
\end{tabular}
\end{table}

We first evaluated the effectiveness of the proposed re-weighting algorithm on the outerwear subset of the database for real users and simulated ideal users. Both experiments were repeated $124$ times, and the average feature weights were calculated for the two experimental conditions. Moreover, we performed $124$ rounds of experiments based on individual features with the ideal user and calculated the average number of search iterations for each feature as an indicator of the discriminative power of that feature. Table \ref{tab::weight} compares the average feature weights for real users and simulated ideal users. As shown in the table, the {\em Style} and {\em Button Shape} features have more discriminative power than  the {\em Category}, {\em Season}, and {\em Color} features.
For the ideal user, the proposed algorithm is able to assign higher weights to the {\em Style} and {\em Button Shape} features and lower weights to the {\em Season} and {\em Color} features.
Meanwhile, for real users, the {\em Color} features has a high weight, whereas the other features have relatively equal weights. The weight of {\em Color} is higher for real users than for ideal users, which may be attributed to the fact that real users tend to focus more on eye-catching features such as color. The weights of the ideal and real users represent the ``system metric'' and ``user metric'', respectively. The goal of our algorithm is to model the high-level user metric based on relevance feedback. %The less average search iteration of the proposed algorithm than the individual feature on simulated ideal user proves the effectiveness of the proposed re-weighting algorithm.

\begin{figure}[t]
\centering
\includegraphics[width=.47\textwidth]{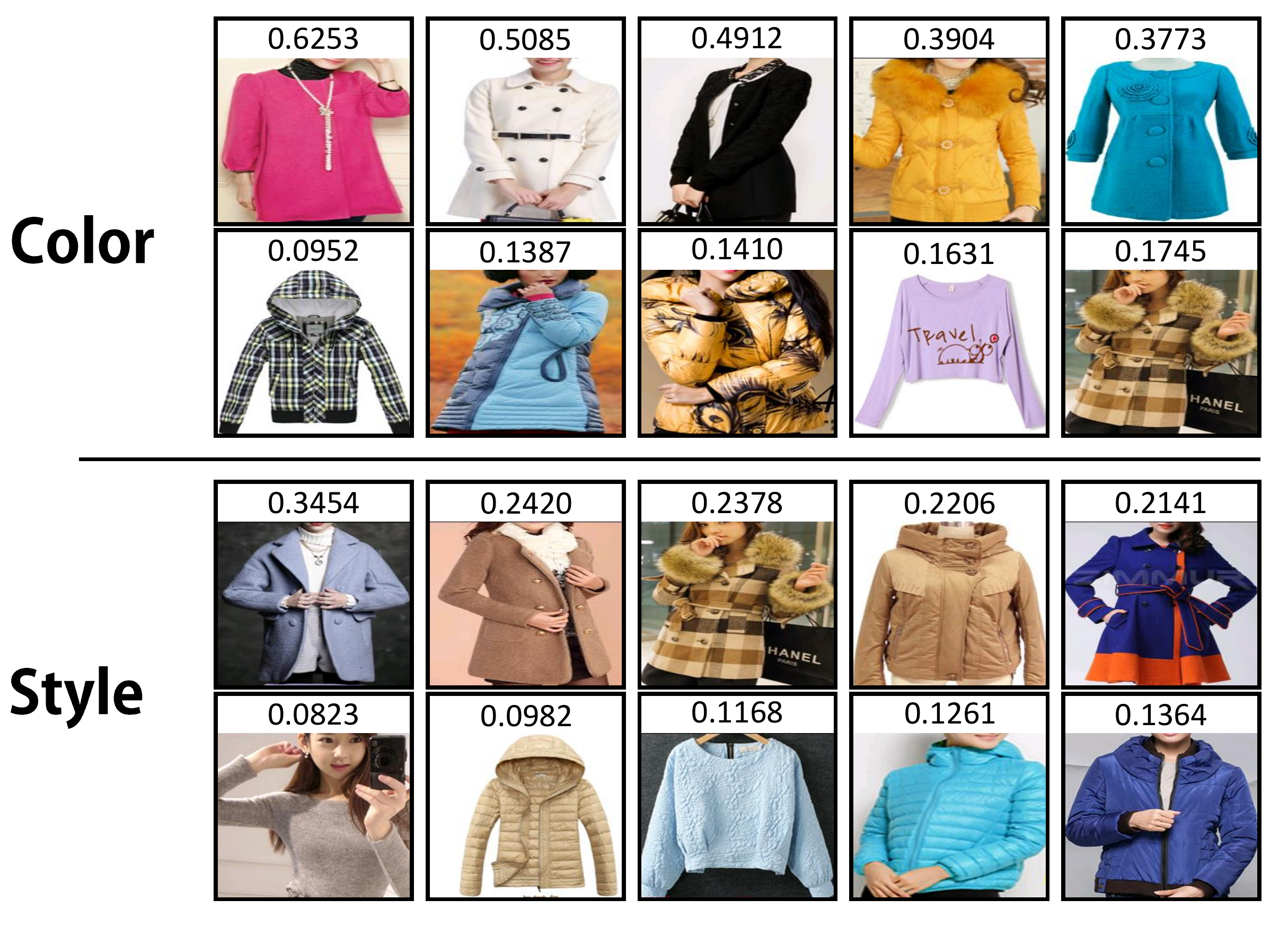}
\caption{Images with the highest and lowest weights for the color and style attributes according to user studies.}
%Most of the images that have a lower weight on colors are relatively confusing to be described by pure color
\label{fig::bias}
\end{figure}

Figure \ref{fig::bias} shows the images with the highest and lowest weights for the {\em Color} and {\em Style} features according to user studies. Query images with prominent colors or with inconspicuous properties related to other features usually result in a higher weight for the color attribute. Similarly, images that are difficult to describe based on patterns receive lower weights in terms of style.

\subsubsection{Scaling Ability}
Figure \ref{fig::scale} shows the average number of search iterations as the number of images in the database increases. The results show that the average number of iterations is approximately linear with respect to the logarithm of the size of the database. Considering that human annotators tend to lose patience after $10$ or $15$ iterations and real users need approximately $1.5$ times the iterations required by the ideal user to complete a search session, we suggest that for target search applications, the size of the database should not exceed $2000$ images.

Several recent studies have attempted to scale up mental image retrieval applications \cite{ferecatu2007interactive, suditu2011heat, suditu2012iterative}. In fact, prior to \cite{fang2005experiments}, such algorithms were mainly extended by applying pre-defined clustering for categorical searches and by applying a tree-and-trace approach for Voronoi partitioning,
%%%Editor - Please ensure that the intended meaning has been maintained
%%%in the above edit.
both of which are complementary to our algorithm. Therefore, our algorithm can be scaled up following similar procedures.

\subsubsection{Consistency}
It can be argued that in addition to the effects of different users and different target images, the choice of the initially displayed images can also have an impact on a search session. We present a further user study conducted to analyze the consistency among the weight distributions generated in different search sessions for the same target image. Using the same target image, for each session, we randomly selected $8$ images to display in the first iteration to initialize the search. The experiment was performed using $10$ target images, and $5$ individual search sessions were conducted for each. Figure \ref{fig::weight} depicts the mean and standard deviation of the numbers of iterations required in the mental image search sessions.
%Table \ref{tab::stdweight} shows the comparison between the standard deviation of feature weights using random target images and the same target image.
It is clear that users refer to similar criteria when searching for the same target image, which is consistent with the assumption that the ``user metric'' is related to the characteristics of the target image.

\section{Conclusion}
\label{sec::conclusion}
In this paper, we investigate a new form of clothing retrieval problem in which an image of the target item resides only in the user's mind. Because of the absence of an explicit image to use as a query, we propose a new Bayesian framework based on implicit relevance feedback for query-free image retrieval. Our algorithm consecutively updates the posterior probability distribution of the target and the weights of multiple features according to user feedback. Based on heterogeneous features extracted from clothing attributes using deep CNNs, a significant advantage of our search-dependent re-weighting scheme is that it models the variability of human decision-making through implicit feedback. Experimental results show that the proposed algorithm consistently outperforms previously developed algorithms based on a pre-defined image similarity metric. As an active attempt to model the subjective nature of user's retrieval needs with limited user interaction, our algorithm also has possible applications in image retrieval and management tasks performed on personal cellphones or community-based media sharing websites.

\ifCLASSOPTIONcompsoc
\section*{Acknowledgments}
\else
\section*{Acknowledgment}
\fi This work is supported by The High Technology Research and
Development Program of China (2015AA015801), NSFC (61221001), and
STCSM (12DZ2272600).

\ifCLASSOPTIONcaptionsoff
  \newpage
\fi

\bibliographystyle{abbrv}
\bibliography{xz3030}

\end{document}